\documentclass[10pt,twocolumn,letterpaper]{article}

\usepackage{cvpr}
\usepackage{times}
\usepackage{epsfig}
\usepackage{graphicx}
\usepackage{amsmath}
\usepackage{amssymb}

\usepackage{booktabs}
\usepackage{ragged2e}
\usepackage{multirow}
\usepackage{float}
\restylefloat{table}

\usepackage[breaklinks=true,bookmarks=false]{hyperref}

\cvprfinalcopy 

\begin{document}

\title{A Neural Temporal Model for Human Motion Prediction}

\author{Anand Gopalakrishnan$^1$, Ankur Mali$^1$, Dan Kifer$^1$, C. Lee Giles$^1$, Alexander G. Ororbia$^2$ \\
Pennsylvania State University, University Park, PA, 16801$^1$ \\
Rochester Institute of Technology, Rochester, NY, 14623$^2$ \\ 
{\tt\small \{aug440, aam35, duk17, clg20\} @psu.edu$^1$, ago@cs.rit.edu$^2$}}

\maketitle

\begin{abstract}
We propose novel neural temporal models for predicting and synthesizing human motion, achieving state-of-the-art in modeling long-term motion trajectories while being competitive with prior work in short-term prediction and requiring significantly less computation. Key aspects of our proposed system include: 1) a novel, two-level processing architecture that aids in generating planned trajectories, 2) a simple set of easily computable features that integrate derivative information, and 3) a novel multi-objective loss function that helps the model to slowly progress from simple next-step prediction to the harder task of multi-step, closed-loop prediction. Our results demonstrate that these innovations improve the modeling of long-term motion trajectories. Finally, we propose a novel metric, called Normalized Power Spectrum Similarity (NPSS), to evaluate the long-term predictive ability of motion synthesis models, complementing the popular mean-squared error (MSE) measure of Euler joint angles over time. We conduct a user study to determine if the proposed NPSS correlates with human evaluation of long-term motion more strongly than MSE and find that it indeed does. We release code and additional results (visualizations) for this paper at:  \textcolor{magenta}{\hyperlink{}{https://github.com/cr7anand/neural\_temporal\_models}}  
\end{abstract}
\section{Introduction}
\label{sec:intro}
We address the problem of building predictive models of human movement using motion capture data. Specifically, we explore models that can successfully be used in forecasting the 3D pose of a human subject conditioned on a small, initial history (a set of priming frames). Current work has focused on two separate but complementary sub-tasks: 1) short-term motion prediction, which is generally evaluated quantitatively by measuring mean squared error (MSE) over a short horizon, and 2) long-term motion prediction, which is evaluated qualitatively by manual, visual inspection of samples in order to evaluate plausible trajectories of human motion over long spans of time. Short-term models are useful in applications of motion tracking while long-term models are useful as motion generation tools for computer graphics \cite{synth_and_track,char_control,motion_graphs}. Models successful in these sub-tasks are also valuable for human gait analysis, studies in the kinematics of human motion, and in human-computer interaction applications \cite{app_kinematics,app_hci}.

Solving the above two sub-problems in motion prediction is challenging given the high dimensionality of the input data as well as the difficulty in capturing the nonlinear dynamics and stochasticity inherent in human motion from observations alone.  Furthermore, human motion, in strong contrast to the motion of other objects, depends on the subject's underlying intent and high-level semantic concepts which are tremendously difficult to model computationally.
Traditionally, models were built in the framework of expert systems and made use of strong simplifying assumptions, such as treating the underlying process as if it was Markovian and smooth or using low-dimensional embeddings \cite{gpdm,linear_human_motion}. These approaches often led to less-than-satisfactory performance. With the modern successes of artificial neural networks \cite{dl_nature} in application domains ranging from computer vision \cite{alexnet} to machine translation \cite{nmt} and language modeling \cite{ororbia2017learning}, many current models of motion are becoming increasingly based on neural architectures.

In this paper, we attack the above two sub-problems using the following strategies. First, we augment the joint angle feature vector, typically fed into predictive neural models, with motion derivative information. This can be easily computed using a finite-difference approximation and naturally contains (temporally) local information that is crucial for generating smooth and consistent motion trajectories. Furthermore, our results demonstrate that the popular approach of training recurrent neural networks (RNNs) in an open loop, i.e., where ground truth input data is fed in at every time step $t$ to predict output at $t+1$, is insufficient when using these models for closed-loop test scenarios, i.e., where model output at time step $t$ is itself used as input to model at time step $t+1$. In the case of closed-loop generation, the model fails to make good predictions over long time horizons due to drifting and an accumulation of next-step error. To remedy this, we introduce a simple, novel multi-objective loss function that balances the goals of effective next-step prediction with generating good long-term, closed-loop predictions, which we find greatly alleviates model drifting. The neural architectures we propose, which make use of a novel, differentiable backward-planning mechanism, are computationally less expensive and far simpler than competing alternatives.

Finally, we address the dearth of effective quantitative methods for evaluating long-term motion synthesis by proposing a novel metric we call the Normalized Power Spectrum Similarity (NPSS). NPSS is meant to complement the popular MSE by addressing some of its drawbacks including: a) a frequency-shift in a predicted sequence, e.g., walking at a faster or slower rate, compared to ground-truth will be heavily penalized by MSE despite being qualitatively similar, and b) a phase-shift in predicted sequence, e.g., if a few frames of motion are missed/skipped by the model, the resulting predicted motion sequence will be phase-shifted but MSE will heavily penalize it despite qualitative similarity with ground-truth. Our measure accounts for these issues as it is designed to capture the difference in the power spectrum of the ground truth frames and the model's predicted joint angles. The key contributions of this work include: 1) a novel, two-stage processing architecture, 2) augmenting the input space with easily computable features useful for the domain of motion, 3) development of a novel loss function that can help guide the model towards generating longer-term motion trajectories and, 4) a novel, evaluation metric called NPSS for long-term human motion quality evaluation, which we will validate with a human user study.

\section{Related Work}
\label{sec:lit-review}
Research in motion synthesis has a long history, with many models proposed over the years. Only in recent times have neural architectures come to the forefront of this domain, quickly supplanting classical statistical learning approaches and hand-crafted methods.
\cite{frag} proposed two architectures: 1) the LSTM-3LR  and 2) the ERD (Encoder-Recurrent Decoder). The LSTM-3LR consists of $3$ layers of $1000$ Long Short-Term Memory (LSTM) units whereas the ERD model uses $2$ layers of $1000$ LSTM units and nonlinear multilayer feedforward networks for encoding and decoding. However, the authors observed that, during inference, the models would quickly diverge and produce unrealistic motion. They alleviate this by gradually adding  noise to the input during training which helps in generating plausible motion over longer horizons. \cite{jain} proposed the  Structural-RNN (SRNN), which takes a manually designed spatio-temporal graph and converts it into a multilayer RNN architecture with body RNNs being assigned to model specific body parts and edge-RNNs to model interactions between body parts. This work also uses the noise scheduling technique employed earlier by \cite{frag} to alleviate drifting. They show that their network outperforms previous methods in both short-term motion prediction as well as long-term qualitative motion. Recently, \cite{jul} proposed simple but hard-to-beat baselines on short-term motion prediction as well as a 1-layer sequence-to-sequence (seq2seq) model \cite{seq2seq} with 1024 Gated Recurrent Unit (GRU) units and a linear output decoder for short-term and long-term motion prediction. Additionally, they trained their long-term model using a sampling-loss as a simpler alternative to noise scheduling in order to alleviate drifting. More recently, \cite{ghosh} proposed a model that couples a denoising autoencoder and an LSTM-3LR network to alleviate drifting for long-term motion synthesis. However, a drawback of their approach is that both the autoencoder and LSTM-3LR networks are first pre-trained independently followed by a subsequent joint fine-tuning step.  

\section{A Neural Motion Synthesizer}
\label{sec:arch}
In this section, we will describe our neural system for motion synthesis, which integrates a novel architecture with a novel loss function and useful, easily computable features. Since our focus is on the specific problem of motion synthesis, we will start by first detailing the benchmark we will test our models against. 

\subsection{Data and Preprocessing}
\label{sec:data}
Staying consistent with previous work on human motion synthesis \cite{frag,jain}, we use the Human 3.6 Million (h3.6m) dataset \cite{h3.6m}, which is currently the largest publicly available motion capture (mocap) database. The h3.6m dataset consists of $7$ actors performing $15$ different actions. Previous work \cite{frag,jain,jul} has been particularly focused on $4$ out of these $15$ categories, e.g., walking, eating, smoking, and discussion when evaluating model performance. To create the test-set, we follow prior work by extracting $8$ motion sequences per action type from subject \#5, yielding the exact same 32 test sequences as used in \cite{frag,jain}. The remaining sequences for subject \#5 are then placed into a validation subset that is used for tuning hyper-parameters. The data of the other six subjects is then used as a training set. We furthermore adopt the pose representation and evaluation metrics as used previously in \cite{frag,jain} to facilitate experimental comparability. Pose is represented as an exponential map of each joint (refer to \cite{gtaylor} for further details). To evaluate our models, we measure the Euclidean distance between predictions and ground truth in Euler angle-space at various time slices along the predicted sequence.

\subsection{Architecture}
\label{sec:arch}
\begin{figure}[!h]
\raggedleft
\includegraphics[width=0.5\textwidth]{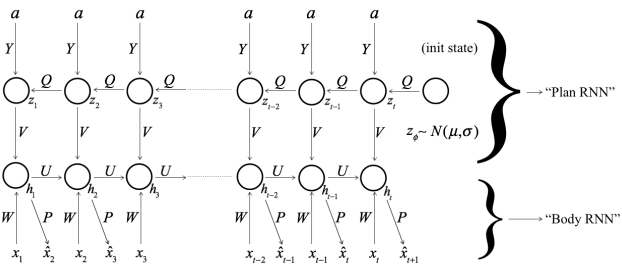}
\caption{VTLN-RNN architecture}
\label{fig:model-diag}
\end{figure}

The architecture we propose for human motion prediction and synthesis we call the Verso-Time Label Noise-RNN model (VTLN-RNN), which consists of a top-level and a bottom-level RNN. Combined, the two RNNs have fewer parameters than prior motion deep learning motion synthesis models. The top-level RNN is meant to serve as a learnable noise process inspired by the work of \cite{ororbiaandmali}, which runs backwards in time, starting from a sampled initial hidden state ($z_{\phi}$) and is conditioned on the one-hot encoding of the action label. This noise process is used to generate a sequence of $K$ ``guide vectors'' (where $K$ is the number of future frames we want to predict, or the prediction horizon) that will be subsequently used by the lower-level RNN. The lower-level RNN, or the Body-RNN, runs forward in time, taking in as input at each time step the joint angle vector $\mathbf{x}_t$ as well as the corresponding guide vector $\mathbf{p}_t$ to generate a prediction of the mocap angles for time-step $t+1$. In essence, running the VTLN-RNN entails using the top-level noise process RNN to generate the guide vectors and then using the Body-RNN to integrate both the bottom-up mocap input vectors and the top-down guide vectors to compute the final hidden states $\mathbf{h}_t$ and the next-step predictions $\mathbf{\hat{x}}_t$. The unrolled model is depicted in Figure \ref{fig:model-diag}. The loss is computed using the Body-RNN's predicted outputs and the corresponding ground-truth mocap vectors.

In order to sample the initial hidden state of the top-level noise process, we first structure it to work like a multivariate Gaussian distribution, drawing inspiration from the re-parameterization trick \cite{kingma} and the adaptive noise scheme proposed in \cite{ororbiaandmali}. The initial state $z_{\phi}$ of the top-level noise process is computed as $z_{\phi} = \mu + \Sigma \otimes \epsilon$, where $\epsilon \sim \mathcal{N} (0, \textrm{I})$, $\mu$ the mean of the random variable, and $\Sigma$ is its covariance, specifically a diagonal covariance. $\mu$ and $\Sigma$ are parameters that are learned along with the rest of the neural network weights using back-propagated gradients during training. This formulation of the hidden state allows the designer to input samples from a simple base distribution, e.g., a standard Gaussian, instead of having to tune the noise parameters, such as its variance, by hand.

In this paper, we use the Gated Recurrent Unit (GRU) \cite{gru} to instantiate both the top-level and bottom-level RNNs of the VTLN-RNN due to its simplicity, competitive performance, and ease of training compared to the LSTM \cite{lstm}. The cell update equations for the top-level GRU remain the same as described in \cite{gru} except that we note its non-state input is the action label (which remains fixed over the length of the sequence). However, the cell update equations for the Body-GRU are as follows:
\begin{align}
r_j = & \sigma \big ( [W_r x_t]_j + [U_r h_{t-1}]_j + [V_r p_{t}]_j \big ) \\
z_j = & \sigma \big ( [W_z x_t]_j + [U_z h_{t-1}]_j + [V_z p_{t}]_j \big ) \\
\tilde{h_j}^t = & \Phi \big ( [W x_t]_j + [U (r \otimes h_{t-1})]_j + [Vp_t]_j \big ) \\
h_j^{t} = & z_j \otimes h_j^{t-1} + (1 - z_j) \otimes \tilde{h_j}^{t}\mbox{.}
\end{align}
The motivation behind the VTLN-RNN structure was to hierarchically decompose the motion synthesis problem into a two-level process, much as has been successfully done in neural-based dialogue modeling \cite{serban2017hierarchical,serban2017piecewise}. The top-level RNN would serve to roughly sketch out a course trajectory that the lower-level RNN would take in, further conditioned on actual data and its own internal state. However, unlike the hierarchical neural dialogue models that served as inspiration, our top-level process runs in the opposite (temporal) direction of the data itself, i.e., backwards. We chose to do this considering gradient flow in the unrolled computation graph. If the top-level process starts at time $t=K$ and works backward to time $t = 1$, the parameter updates of the top-level model will depend more heavily on information from the future (or from far later on in a sequence) and this information would be encoded in the synaptic weights related to a specific action type/label. When the top-level process is used to generate the sequence of guide vectors, it creates ``hints'' or coarsely defined states that the lower-level Body-RNN can then refine based on actual input data or its own closed-loop predictions. 

While it is hard to prove that the top-level RNN is truly ``planning'' out the ultimate trajectory of the model's predictions, our experiments will show that our two-level process offers some useful regularization, improving model generalization over  mechanisms such as drop-out. 
Additionally, our model used for both short-term and long-term motion prediction has significantly fewer parameters compared to \cite{frag,jain,jul} and yet achieves state-of-art results as shown in Table \ref{tab:long-term} on long-term motion prediction. Furthermore, it is competitive with \cite{jul} state-of-art results on short-term motion prediction as shown in Table \ref{tab:short-term}.

\begin{table}[!h]
    \centering
    \setlength{\tabcolsep}{2pt}
    \begin{tabular}{c|c}
        Models & No. of parameters \\
        ERD \cite{frag} & 14,842,054\\
        LSTM-3LR \cite{frag} & 20,282,054\\
        SRNN \cite{jain} & 18,368,534\\
        MBR-long \cite{jul} & 3,425,334\\
        GRU-d (ours) & \textbf{2,735,670}\\
        VGRU-d (ours) & \textbf{3,413,047}\\
    \end{tabular}
    \caption{Number of parameters of models. }
    \label{tab:model-param}
\end{table}

\subsection{Incorporating Derivative Information}
\label{sec:deriv}
Motion derivatives contain crucial feature information used to model local (near past) motion information. These features are cheap to compute and do not require any additional model parameters. Motivated by this, we extract motion derivatives by a using a finite backward difference approximation, calculated as follows:
\begin{equation}
\nabla_{h}^{n} [f] (x) = \sum_{i=0}^{n} (-1)^{n} { n \choose i } f(x - ih)
\end{equation}
where $i$ indexes the order of the derivative we would like to approximate, up to $n$, and $h$ is a non-zero spacing constant. \\
We extract the $n = \{1,2,3\}$ motion derivatives with $h=1$ using the above equation and append these vectors to the vector of joint angles. The linear decoder of our recurrent model outputs only joint angles for the next time step. During closed loop, iterative multi-step prediction we calculate these motion derivatives on-the-fly.

\subsection{Facilitating Closed-Loop Prediction} 
\label{sec:multi-loss}
The standard way to train RNNs for sequence prediction tasks is to feed the ground truth inputs at every time step during training. Then, at test time, the model's previous prediction at $t$ is fed in, treating it as it were ground truth input, when making a prediction at $t+1$. This process is known as closed-loop (or iterative) prediction. However, a key issue with this method is that the model is unable to recover from the accumulation of errors leading to a significant degradation in the RNN's predictions over time. This is due to the strong mismatch in the inputs presented during training (i.e. ground-truth inputs) and test time (i.e. the model's own noisy predictions from previous time steps). This causes synthesized long-term motion trajectories to quickly diverge from the manifold of plausible motion trajectories. As mentioned earlier, \cite{frag} and \cite{jain} alleviate this issue by injecting Gaussian noise, of gradually increasing magnitude, to inputs during training. \cite{jul} used a sampling loss where, during training, the model outputs are fed back to itself. Professor Forcing \cite{prof_force} addresses this issue by using an adversarial training regime to ensure the hidden states of the RNN are constrained to be similar during train and test time. However, this method is computationally expensive, needs careful hyperparameter tuning, and suffers from stability issues normally encountered in the training of Generative Adversarial Networks. More recently, \cite{auto_cond} showed that their method, or auto-conditioning, helps the RNN models produce good qualitative long-term motion by alternating between feeding in ground-truth samples and the model's own outputs during training.

We view the problem of using the RNN for multi-step iterative prediction at test time from the perspective of multi-task and curriculum learning. We ultimately require the RNNs to achieve good performance on the hard-task of multi-step iterative prediction starting from the simple task of one-step prediction. An intuitive way to achieve this would be to gradually make the RNN progress from the simple task of one-step prediction (ground truth fed in at every time step) to the final goal of multi-step iterative prediction. Defining a composite loss function with separate terms for measuring one-step prediction and multi-step iterative prediction losses, and weighting these terms, would ensure that the network slowly adapts from being able to only predict one-step ahead to becoming capable of multi-step iterative prediction during the course of the training cycle. This intuition forms the basis of our multi-objective loss function defined as follows,
\begin{equation}
L(\widehat{y}, y) = \frac{1}{T} \sum_{t=0}^{T} (\widehat{y_{o}}^{t} - y^t)^2 + \frac{\lambda}{T'} \sum_{t_1 = 0}^{T'} (\widehat{y_{c}}^{t_1} - y^{t_1})^2
\label{eqn:dan-loss}
\end{equation}
where $y^t$ = ground-truth output at $t$, $\widehat{y_{o}}^{t}$ = model output in open-loop mode at $t$, $\widehat{y_{c}}^{t_1}$ = model output in closed-loop mode at $t_1$. Open-loop mode refers to feeding ground-truth inputs at every time step to the RNN in order to produce outputs and closed-loop mode refers to feeding the model's own output at $t$ as input to it at $t+1$. For every input sequence, this loss requires us to run the forward pass twice: 1) compute  $\widehat{y_{o}}^{t}$ in open-loop mode, and 2) compute  $\widehat{y_{c}}^{t}$ in closed-loop mode. We gradually increase $\lambda$ using a step schedule over the training cycle starting from zero at the beginning. This schedule therefore gradually places greater importance on the loss-term focused on closed-loop predictions as the network learns to make better one-step predictions. In our long-term motion synthesis experiments, we will see that our loss outperforms noise scheduling \cite{frag,jain}, auto-conditioning \cite{auto_cond}, and the sampling loss of \cite{jul}.

\subsection{A Complementary Long-Term Motion Metric}
\label{sec:eval-metrics}
The use of mean-squared error (MSE) as an evaluation metric for models has been standard practice \cite{frag,jain,jul} for both short-term motion prediction and long-term motion synthesis tasks. In short-term motion prediction, the metric needs to capture how well a model mimics ground-truth data over short-term horizons (i.e 0-500 milliseconds) since it is to be used in motion tracking applications.

However in the long-term motion synthesis task, models need to be evaluated on how well they generate plausible future motion over long-term horizons given some seed frames of motion. Since human motion is inherently stochastic over long time horizons, models can significantly deviate from the ground-truth trajectories and have a large MSE despite producing qualitatively good human motion. This problem has been noted in prior work \cite{frag,jain,jul}. There are a variety of causes. For example, if the predictions correspond to walking at a slower pace, the joint angles will be misaligned (frequency-shift) and MSE computed will diverge over time. In the short term, the joint angles may still be similar enough for MSE to meaningfully capture similarity, but in the long term they will become significantly different. Similarly, adding or removing a few extra frames of motion (phase-shift) compared to ground-truth sequence will result in high MSE values because frames are, again, misaligned. Therefore, as noted in prior work \cite{frag,jain,jul}, the use of MSE as an evaluation metric is not appropriate for the long-term task. However, no attempt has previously been made to suggest another quantitative metric for evaluating long-term motion synthesis models. 

In this paper, we propose such a metric, backed by a user study, based on the following intuition. We can say that the qualitative essence of any action, e.g., walking, eating, running, can be captured through the frequency signature of joint angles of the body while performing that action. Take walking at a slower pace as an example -- the power spectrum (obtained from a discrete Fourier transform) would show spikes at a slightly lower frequency and the addition or removal of a few frames would show up as a phase-shift in the frequency domain. The examples of slow/fast or phase-shifted walking involve periodic sub-actions, whereas aperiodic actions, such as discussion, will show a more uniform spread of power in the frequency domain (this indicates a lack of periodicity in the action which is also being picked up by the power spectrum). Measuring similarity of the power spectrum between between a ground truth sequence and a corresponding generated sequence for the same motion type would account for these phenomena and correlate better with the visual quality (see user study results in Section \ref{sec:user-study}) of samples. The field of content-based image retrieval have used Earth Mover's Distance (EMD) \cite{tomasi,grauman} to quantify perceptual similarity of images using the EMD between their color histograms. Based on the intuition we have developed so far and the recent successful application of EMD, we propose an EMD-based metric over the power spectrum that overcomes many of the shortcomings of MSE as an evaluation metric on the long-term task.

For a given action class in the test set, let there be $k$ sequences each of $T$ length and output vector of joint angles at each time-step be $D$ dimensional. We define $x_{i, j} [t]$ to be the ground-truth value at time $t$ for $j^{th}$ feature dimension for $i^{th}$ sequence and $y_{i,j} [t]$ to be the corresponding model prediction. Also, let $X_{i,j} [f]$ and $Y_{i,j} [f]$ be the squared magnitude spectrum of Discrete Fourier Transform coefficients (per sequence $i$ per feature dimension $j$ ) of $x_{i, j} [t]$ and $y_{i. j} [t]$ respectively. First we normalize $X_{i,j} [f]$ and $Y_{i,j} [f]$ w.r.t $f$ as,
\begin{align}
X_{i,j}^{\textrm{norm}} [f] = & \frac{X_{i,j} [f]}{\sum_{f} X_{i,j} [f]} ;
Y_{i,j}^{\textrm{norm}} [f] = \frac{Y_{i,j} [f]}{\sum_{f} Y_{i,j} [f]} \\
\textrm{emd}_{i,j} = & \| X_{i, j}^{\textrm{norm}} [f] - Y_{i, j}^{\textrm{norm}} [f]) \|_1 
\label{eqn:emd}
\end{align}
where, $\|.\|_1$ is the L1-norm. Finally, we use a power weighted average over all $i$ and $j$ of 1-D EMD distances computed in (\ref{eqn:emd}) as shown below,
\begin{equation}
NPSS = \frac{\sum_{i} \sum_{j} p_{i, j}*\textrm{emd}_{i,j}}{\sum_{i} \sum_{j} p_{i, j}} \quad
p_{i,j} = \sum_{f} X_{i,j}^{\textrm{norm}} [f] 
\label{NPSS}
\end{equation}
where $p_{i, j}$ = total power of $i^{th}$ feature in the $j^{th}$ sequence, to arrive at our scalar evaluation metric for an evaluation set of sequences for a given action class. We refer to our metric as normalized power spectrum similarity (NPSS). Another interpretation is that we can view long-term motion synthesis as a generative modeling task. By this interpretation, the evaluation metric must capture differences in the distributions of the ground-truth and predicted motion samples. NPSS captures distributional differences in the power spectrum of joint angles of the ground-truth and predicted sequences. As a result, it is better equipped to model differences in the visual quality of motion trajectories. 

\section{Experiments}
\label{sec:experiments}

\subsection{Training Setup}
\label{sec:setup}
For our short-term model, the VGRU-r1 (MA), we trained on all action classes using our proposed multi-objective cost, calculating gradients over mini-batches of $32$ samples (clipping gradient norms to $5$) and optimizing parameters over $100,000$ iterations  RMSprop \cite{rmsprop} with initial learning rate $\lambda = 0.0001$ and decayed by $0.8$ every 5000 iterations until 60,000 iterations. Drop-out \cite{zaremba,pham}, with probability of 0.3, was applied only to the Body-RNN, which was further modified to use skip connections that connect input units to output units, as in \cite{jul}.
The model was given $50$ seed frames and tasked with predicting the next $10$ subsequent frames ($400$ milliseconds). When training for this, the VTLN-RNN is unrolled backwards while the Body-RNN is unrolled forwards, in time, over $60$ steps. (Note: MA stands for multi-action, SA for single-action.)

For our long-term models, which were trained on single-action data, parameter optimization was carried out with RMSprop ($\lambda = 0.0002$, decayed by $0.6$ every $2000$ iterations) over $10,000$ iterations with mini-batches of 32, using, again, our proposed cost function. Models were fed in $50$ seed frames and made to predict the next $100$ frames ($4$ sec), which meant that the VTLN-RNN was unrolled backwards and the Body-RNN forwards $150$ steps. The input vector to the Body-RNN consisted of joint angles appended with motion derivatives. 
VGRU-d refers to our proposed VTLN-RNN architecture where the VTLN-RNN and Body-RNN both contain only a single layer of 512 GRU cells. GRU-d refers to a 2-layer GRU model (512 units in each). Both VGRU-d and GRU-d models are trained with our proposed loss and make use of inputs augmented with motion derivatives. VGRU-ac refers to our VTLN-RNN architecture trained with auto-conditioning \cite{auto_cond}, using the recommended length of $5$, serving as a baseline.
For all models (short and long-term), hyper-parameters were tuned on a separate validation set.
\begin{figure}[!h]
\centering
\includegraphics[trim=2cm 10cm 2cm 8cm,clip,width=0.5\textwidth]{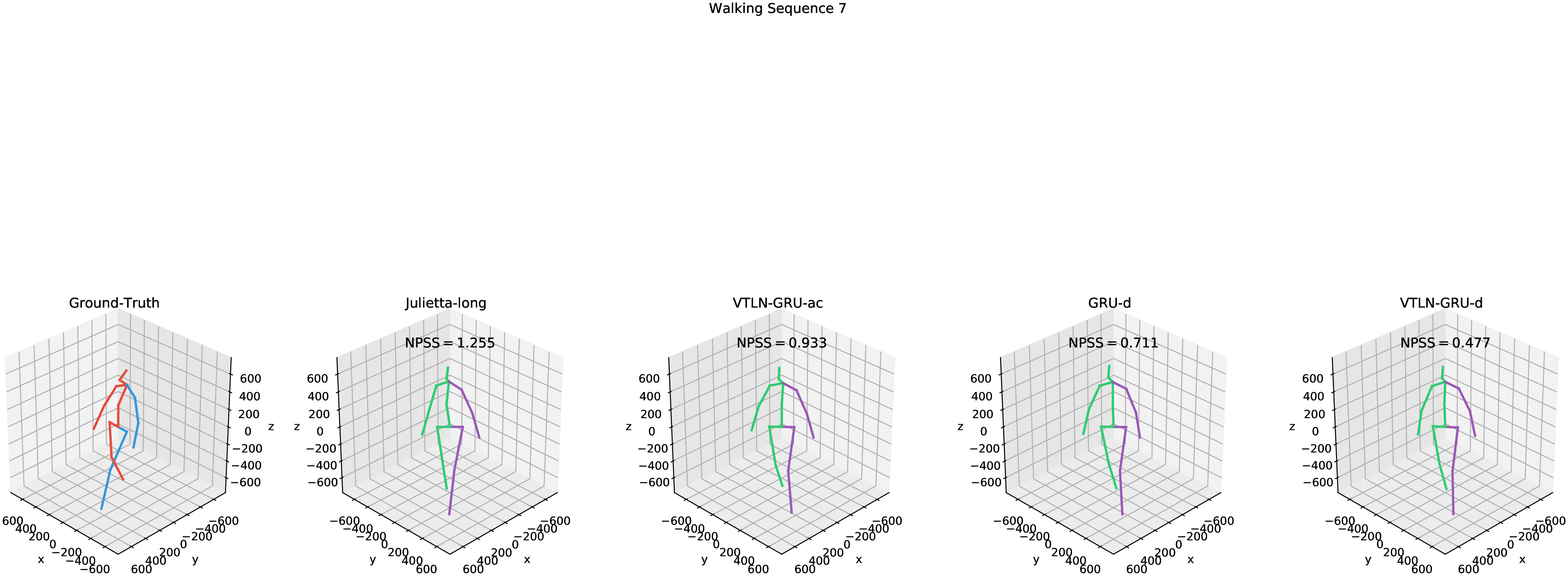} \\
\includegraphics[trim=2cm 10cm 2cm 8cm,clip,width=0.5\textwidth]{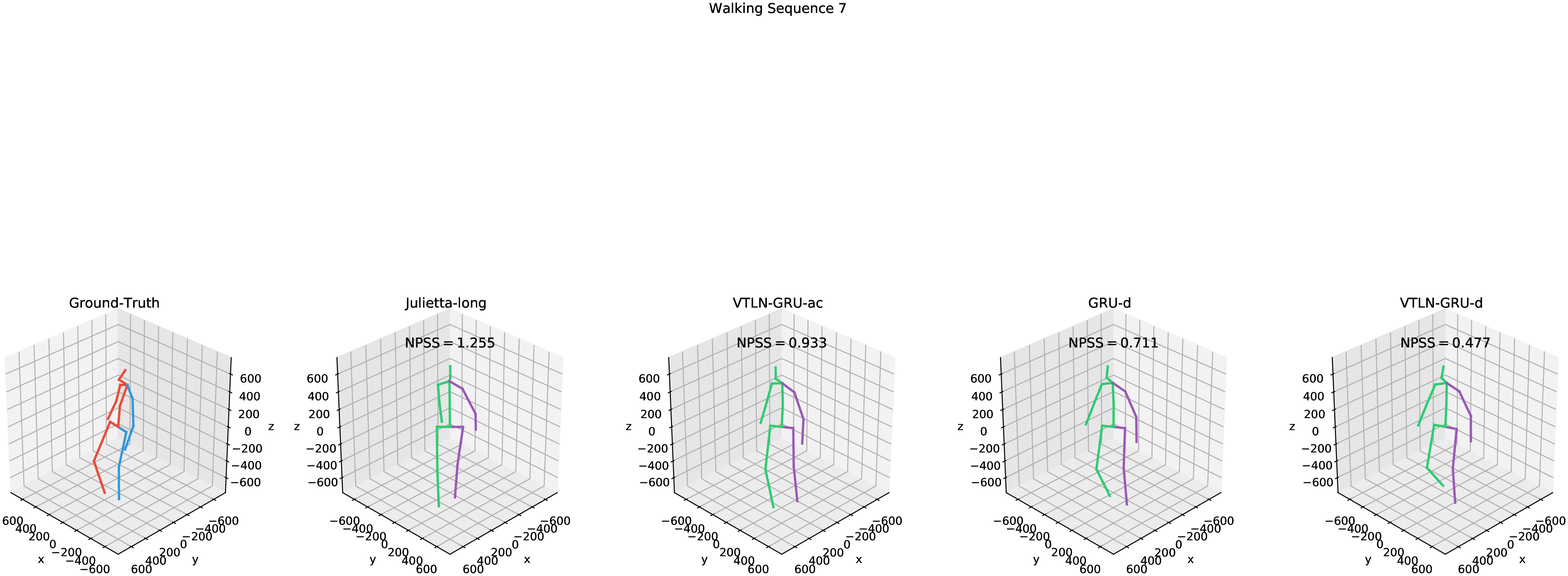} \\
\includegraphics[trim=2cm 10cm 2cm 8cm,clip,width=0.5\textwidth]{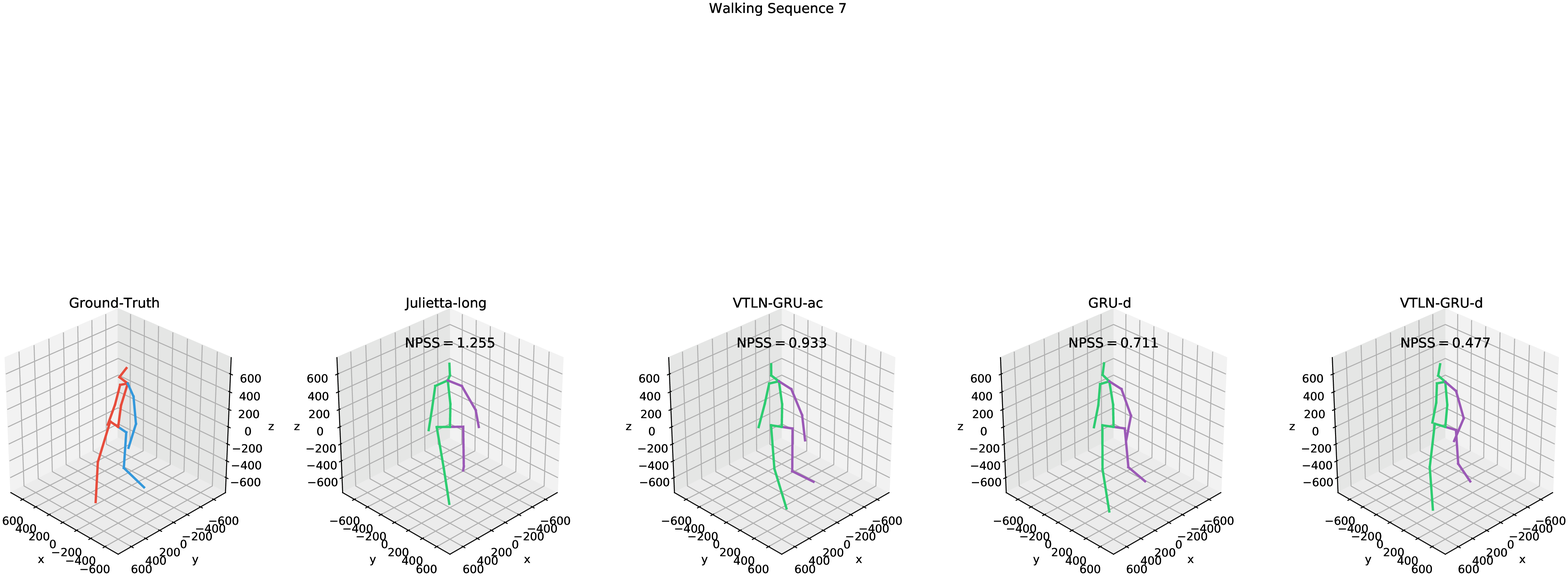} \\
\includegraphics[trim=2cm 10cm 2cm 8cm,clip,width=0.5\textwidth]{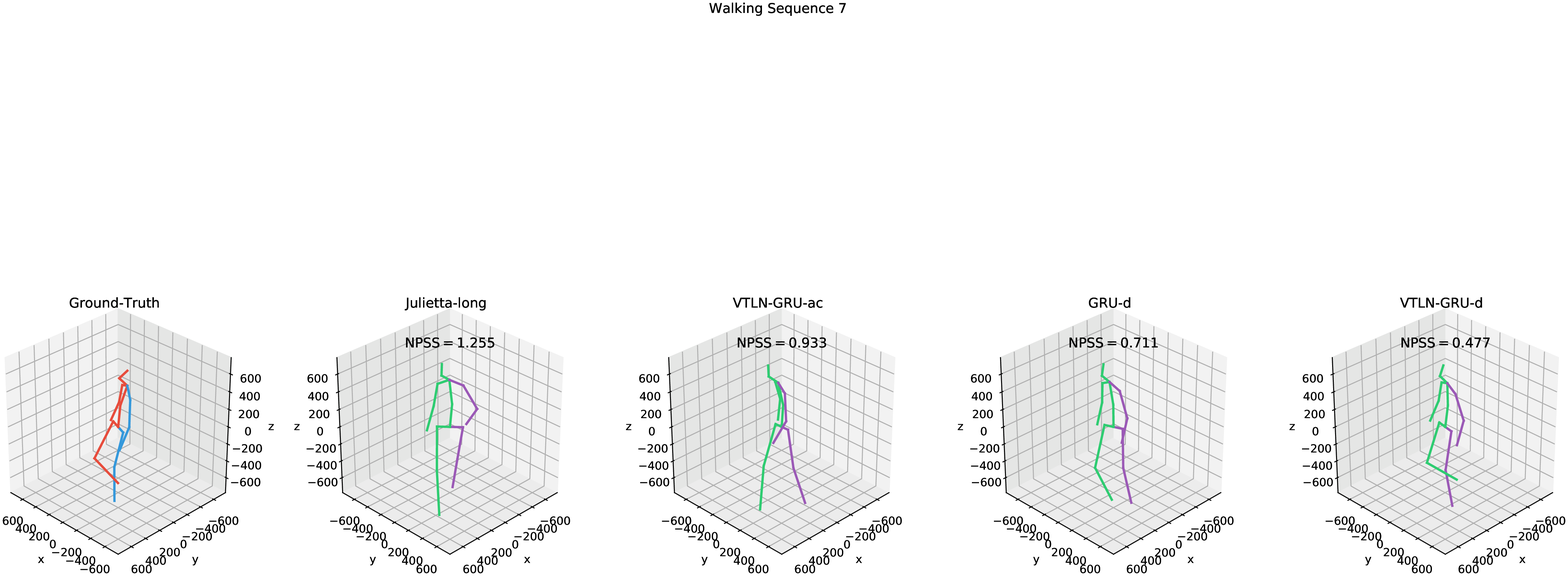} \\
\includegraphics[trim=2cm 10cm 2cm 8cm,clip,width=0.5\textwidth]{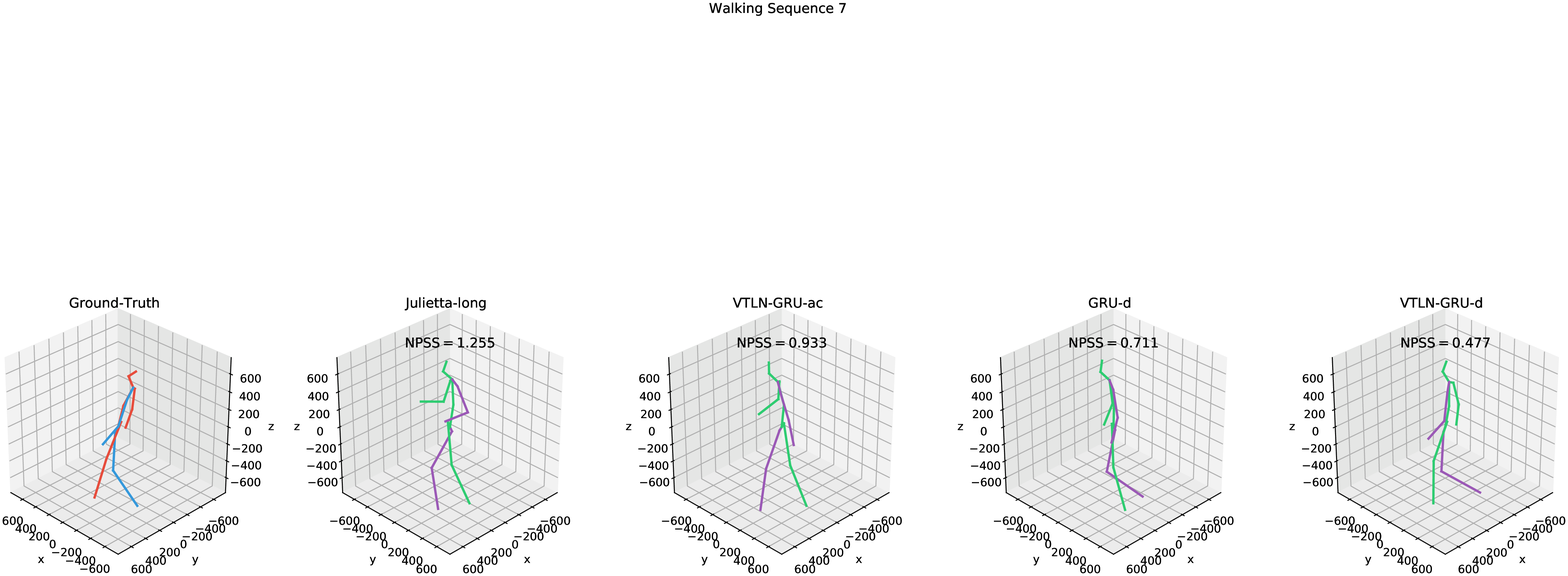}
\caption{Long-term motion synthesis on walking activity on test sequence. Snapshots are shown at 160, 560, 1000, 2000 and 4000 milliseconds (from top-to-bottom) along the prediction time-axis. We see that GRU-d and VTLN-GRU-d are qualitatively closer to the ground-truth sequence than MBR-long and VTLN-GRU-ac.}
\end{figure}

\subsection{User Study: Long-Term Motion Synthesis}
\label{sec:user-study}
We conducted a user study to understand how human judgment of long-term motion correlates with MSE as well as our proposed NPSS. A desirable quantitative evaluation metric for long-term human motion would be one that strongly agrees with human judgment. In order to conduct this study, we considered the 6 models from Table \ref{tab:npss-full} (i.e. VGRU-r1(SA), MBR-unsup (SA), MBR-long, VGRU-ac, GRU-d and VGRU-d). In each trial, a user was presented videos of the ground-truth motion and corresponding model predictions from a randomly chosen pair of models (from the list above) for a given test-set action sequence (the ordering of the models was random and identities were hidden from the users). Users were asked to compare model-generated motion trajectories with the ground truth, based on which one possessed better ``motion quality''. The users were informed that the phrase ``motion quality'' referred to similarity/closeness in overall skeletal pose (i.e. overall posture) and joint motion dynamics over the entire sequence, rather than simple point-to-point matches in time, and made their decisions based on this criteria. Please refer to the supplementary material for a sample screenshot of the user survey video. 

For each of the $4$ action classes (i.e. walking, eating, smoking, and discussion) we presented $20$ video sequences of the ground-truth with the A versus B comparison (refer to supplementary for a sample user study screenshot). Video samples were selected uniformly and randomly (without replacement) from all possible, pairwise combinations of the $6$ models. We then selected a test sequence for an action class, via uniform random sampling with replacement, and presented the ground-truth motion sequence and previously picked paired model predictions for that sequence. This process is repeated to generate 20 videos (i.e. 20 questions) for each of the $4$ actions. The study involved 20 participants for each of the 4 action class surveys.

Now for the 2 evaluation metrics (i.e. MSE and NPSS) we derive rankings of the models used in the user study. For the MSE metric ranking, we compute the sum of MSE over all time slices for the long-term window (i.e. 80, 160, 320, 400, 560, 1000 milliseconds, which is consistent with prior work \cite{jain}). For NPSS, we use the results in Table \ref{tab:npss-full} to arrive at a ranking of the models for all $4$ actions. 
Then, we use these rankings (MSE and NPSS) for each action class to make predictions for each question in the user survey. As shown in Table~\ref{tab:agree}, we compute the probabilities of agreement and disagreement with users for MSE and NPSS.

\begin{table}[!h]
\centering
\resizebox{\columnwidth}{!}{
\begin{tabular}{c|c|c|c|c}
\hline
Metrics & Walking & Eating & Smoking & Discussion \\
\hline
\multirow{6}{*}{MSE rankings} & 1. VGRU-r1(SA) & 1. VGRU-r1(SA) & 1. VGRU-r1(SA) & 1. VGRU-r1(SA)\\
 & 2. MBR-unsup (SA) & 2. VGRU-d & 2. MBR-unsup (SA) & 2. VGRU-d \\
 & 3. VGRU-d & 3. MBR-unsup (SA) & 3. VGRU-d & 3. GRU-d \\
 & 4. MBR-long & 4. VGRU-ac & 4. VGRU-ac & 4. VGRU-ac \\
 & 5. VGRU-ac & 5. GRU-d & 5. GRU-d & 5. MBR-unsup (SA) \\
 & 6. GRU-d & 6. MBR-long & 6. MBR-long & 6. MBR-long \\ 
\hline
\multirow{6}{*}{NPSS rankings} & 1. VGRU-d & 1. GRU-d & 1. VGRU-d & 1. VGRU-ac \\
 & 2. GRU-d & 2. VGRU-ac & 2. GRU-d (SA) & 2. GRU-d \\
 & 3. VGRU-ac & 3. VGRU-d (SA) & 3. VGRU-ac & 3. VGRU-d \\
 & 4. VGRU-r1 (SA) & 4. VGRU-r1 (SA) & 4. VGRU-r1 (SA) & 4. MBR-unsup (SA) \\
 & 5. MBR-long & 5. MBR-unsup (SA) & 5. MBR-unsup (SA) & 5. MBR-long \\
 & 6. MBR-unsup (SA) & 6. MBR-long & 6. MBR-long & 6. VGRU-r1 (SA) \\
\hline 
\end{tabular}}
\vspace{1mm}
\caption{Long-term motion model MSE \& NPSS rankings.}
\label{tab:ranking}
\end{table}


\begin{table}[h]
\centering
\begin{tabular}{|c|c|c|}
\hline
 & MSE & NPSS \\
\hline
Agree & 0.4875 (39/80) & 0.8125 (65/80) \\
Disagree & 0.5125 (41/80) & 0.1875 (15/80) \\
\hline
\end{tabular}
\vspace{1mm}
\caption{User agreement ratios for MSE \& NPSS aggregated across all actions taking majority user vote as ground-truth. a/b = number of times user answers agrees with metric's answer/ total equences in user survey across 4 actions.}
\label{tab:agree}
\end{table}

Furthermore, we conducted a Binomial test of proportions to test the claim that NPSS agrees better with user judgment than MSE. In this test, $p_1$ is defined to be the probability that, on a random sample, NPSS will agree with human ordering/choice while $p_2$ is the probability that MSE will agree with human ordering/choice. We take the null hypothesis to be $H_0: p_1 \leq p_2$ and the alternative hypothesis to be $H_A: p_1 > p_2$ and seek to test the null against the alternative hypothesis. Scientific studies typically set the threshold of statistical significance for p-values to be below $0.01$ (smaller p-values would better support the claim that NPSS is a better metric, confirming that $p_1$ is statistically larger than $p_2$). The value we obtained is significantly lower than this threshold, i.e., a p-value of $\bf{1.7} \times \bf{10^{-5}}$.

\section{Results and Discussion}
\label{sec:discussion}
Given the results of our user study, we argue that NPSS should be preferred (over MSE) for measuring model generation quality over long sequences (for predictions made over longer horizons). However, to holistically evaluate a motion synthesis model, we recommend using NPSS in tandem with MSE when evaluating a model's ability to make both short-term and long-term predictions. The results of our user study for NPSS is promising, however, further studies should be conducted to further validate and strengthen our findings.

For compatibility with prior work, Table \ref{tab:long-term} compares the MSE of Euler angles, measured at particular time slices on test sequences, reporting results for competing methods such as LSTM-3LR and ERD \cite{frag}, SRNN by \cite{jain}, and MBR-long \cite{jul}. Although our short-term model, VGRU-r1 (SA), displays the best performance (lowest mse) until the $1$ second mark, it has been noted by \cite{jain,jul}, and further corroborated by the results of our user study, that MSE is not appropriate for the task of long-term motion synthesis. Table \ref{tab:npss-full} shows the NPSS metric results for models evaluated on the test set. 

\begin{table}[!ht]
\centering
\resizebox{\columnwidth}{!}{%
\begin{tabular}{c|c|c|c|c}
\hline
Models & Walking & Eating & Smoking & Discussion \\
\hline
VGRU-r1(SA) (ours) & 1.217 & 1.312 & 1.736 & 4.884\\
MBR-unsup (SA) \cite{jul} & 1.809 & 1.481 & 2.794 & 2.258\\
\hline
\hline
MBR-long \cite{jul} & 1.499 & 1.621 & 4.741 & 2.882\\
VGRU-ac & 1.032 & 0.842 & 1.426 & \textbf{1.651} \\
GRU-d (ours)& 0.931 & \textbf{0.836} & \underline{1.274} & \underline{1.688}\\
VGRU-d (ours)& \textbf{0.887} & \underline{0.846} & \textbf{1.235} & 1.777\\
\hline
\end{tabular}%
}
\vspace{1mm}
\caption{Test-set NPSS scores (lower is better). Above the double line: short-term models, i.e., MBR-unsup (SA),  MBR-unsup. (MA) \cite{jul} (re-trained on single-action), and ours, sampled for long-term durations. Below the line: long-term models, i.e., MBR-long (SA) \cite{jul}, and ours, such as GRU-d, VGRU-d, \& VGRU-ac.}
\label{tab:npss-full}
\end{table} 

\begin{table}[!h]
\centering
\resizebox{\columnwidth}{!}{%
\begin{tabular}{c|c|c|c|c}
\hline
Models & \multicolumn{4}{|c}{Short-Term} \\
\hline
 & Walking & Eating & Smoking & Discussion \\
\hline
VGRU-r1 (SA) (ours)& 0.120 & \textbf{0.091} & \textbf{0.052} & 0.258 \\
MBR-unsup (SA) \cite{jul} & 0.238 & 0.249 & 0.183 & 0.416 \\
\hline
\hline
MBR-long \cite{jul}& 0.161 & 0.214 & 0.265 & 0.703 \\
VGRU-ac & 0.118 & 0.113 & 0.075 & 0.256 \\
GRU-d (ours)& 0.127 & 0.095 & 0.126 & \textbf{0.185} \\
VGRU-d (ours)& \textbf{0.117} & 0.121 & 0.084 & 0.194 \\
\hline
 & \multicolumn{4}{|c}{Medium-Term} \\
\hline
VGRU-r1 (ours)(SA) & 0.194 & 0.093 & 0.079 & 0.375 \\
MBR-unsup (SA) \cite{jul}& 0.206 & 0.178 & 0.237 & 0.439 \\
\hline
\hline
MBR-long \cite{jul}& 0.237 & 0.160 & 0.405 & 0.477 \\
VGRU-ac & 0.188 & 0.103 & 0.097 & 0.298 \\
GRU-d (ours)& \textbf{0.170} & 0.096 & 0.083 & \textbf{0.258} \\
VGRU-d (ours)& 0.179 & \textbf{0.080} & \textbf{0.067} & 0.331 \\
\hline
& \multicolumn{4}{|c}{Long-Term} \\
\hline
VGRU-r1 (SA) (ours) & 0.544 & 0.764 & 0.948 & 2.72 \\
MBR-unsup (SA) \cite{jul}& 0.884 & 0.684 & 1.077 & 0.943 \\
\hline
\hline
MBR-long \cite{jul}& 0.549 & 0.754 & 1.403 & 1.245 \\
VGRU-ac & 0.460 & 0.459 & 1.051 & 0.811 \\
GRU-d (ours) & 0.406 & 0.332 & 0.723 & \textbf{0.785} \\
VGRU-d (ours) & \textbf{0.359} & \textbf{0.288} & \textbf{0.577} & 1.001 \\
\hline
\end{tabular}%
}
\vspace{1mm}
\caption{NPSS at 3 different time scales i.e 1) short-term: 0-1 second 2) medium-term: 1-2 seconds 3) long-term: 2-4 seconds window prediction on test set}
\label{tab:npss-timescale}
\end{table} 

We can see that the short-term models, VGRU-r1 (SA) and MBR-unsup (SA), despite having the lowest MSE values (until 1 second, as in Table \ref{tab:long-term}), achieve worse scores in terms of NPSS when compared to long-term models. This result is in accordance with the visual quality of samples produced by these models and illustrates how NPSS is better equipped to capture differences in sample quality than MSE. 
Based on the NPSS metric, VGRU-d and GRU-d produce better long-term motion trajectories, outperforming MBR-long and VGRU-ac across all $4$ action classes.

\begin{table*}[t]
\centering
\small
\tabcolsep=0.7mm
\resizebox{\textwidth}{!}{\begin{tabular}{@{}lrrrr|rrrr|rrrr|rrrr@{}} & \multicolumn{4}{c}{Walking} & \multicolumn{4}{c}{Eating} & \multicolumn{4}{c}{Smoking} & \multicolumn{4}{c}{Discussion}\\
milliseconds & 80 & 160 & 320 & 400 & 80 & 160 & 320 & 400 & 80 & 160 & 320 & 400 & 80 & 160 & 320 & 400 \\
\midrule
Zero-velocity \cite{jul} & 0.39 & 0.68 & 0.99 & 1.15 & 0.27 & 0.48 & 0.73 & 0.86 & \textbf{0.26} & \textbf{0.48} & 0.97 & 0.95 & \textbf{0.31} & \textbf{0.67} & \textbf{0.94} & \textbf{1.04} \\
MBR-unsup (MA) \cite{jul} & \textbf{0.27} & \textbf{0.47} & 0.70 & 0.78 & 0.25 & 0.43 & 0.71 & 0.87 & 0.33 & 0.61 & 1.04 & 1.19 & 0.31 & 0.69 & 1.03 & 1.12 \\
MBR-sup (MA) & 0.28 & 0.49 & 0.72 & 0.81 & \textbf{0.23} & \textbf{0.39} & \textbf{0.62} & \textbf{0.76} & 0.33 & 0.61 & 1.05 & 1.15 & 0.31 & \underline{0.68} & 1.01 & 1.09 \\
VGRU-r1 (MA) (ours) & 0.34 & \textbf{0.47} & \textbf{0.64} & \textbf{0.72} & \underline{0.27} & \underline{0.40} & \underline{0.64} & \underline{0.79} & 0.36 & \underline{0.61} & \textbf{0.85} & \textbf{0.92} & 0.46 & 0.82 & \underline{0.95} & 1.21 \\
& $\pm$ 1e-3 & $\pm$ 1e-3 & $\pm$ 2e-3 & $\pm$ 2e-3 & $\pm$ 2e-3 & $\pm$ 1e-3 & $\pm$ 2e-3 & $\pm$ 2e-3 & $\pm$ 6e-4 & $\pm$ 1e-3 & $\pm$ 1e-3 & $\pm$ 1e-3 & $\pm$ 2e-3 & $\pm$ 1e-3 & $\pm$ 3e-3 & $\pm$ 5e-3 \\
\bottomrule
\end{tabular}}
\vspace{1mm}
\caption{Short-term results: MSE on test sequences for short-term motion
prediction. All models are trained on multiple actions. VGRU-r1 (MA) refers to our VTLN-RNN with 1 layer (512 GRU unit) and a Body-RNN with 1 layer 512 GRU cells, where Body-RNN has residual input-to-output connections as in \cite{jul}. For the VGRU-r1, model we compute mean and standard error over $30$ trials.}
\label{tab:short-term}
\end{table*}

\begin{table*}[h]
\centering
\tabcolsep=1.0mm
\resizebox{\textwidth}{!}{\begin{tabular}{@{}lrrrrrr|rrrrrr|rrrrrr|rrrrrr@{}}
 & \multicolumn{6}{c}{Walking} & \multicolumn{6}{c}{Eating} & \multicolumn{6}{c}{Smoking} & \multicolumn{6}{c}{Discussion}\\
models & 80 & 160 & 320 & 400 & 560 & 1000 & 80 & 160 & 320 & 400 & 560 & 1000 & 80 & 160 & 320 & 400 & 560 & 1000 & 80 & 160 & 320 & 400 & 560 & 1000 \\
\midrule
MBR-unsup (SA) \cite{jul} & \textbf{0.37} & 0.655 & 0.987 & 1.095 & 1.286 & 1.476 & 0.411 & 0.781 & 1.375 & 1.630 & 1.926 & 2.106 & 0.472 & 0.891 & 1.497 & 1.726 & 2.077 & 2.581 & 0.701 & 1.326 & 2.134 & 2.433 & 2.996 & 2.950\\
VGRU-r1 (SA) (ours)  & 0.410 & \textbf{0.570} & \textbf{0.807} & \textbf{0.868} & \textbf{1.026} & \textbf{1.231} & \textbf{0.285} & \textbf{0.441} & \textbf{0.668} & \textbf{0.829} & \textbf{0.995} & \textbf{1.531} & \textbf{0.378} & \textbf{0.656} & \textbf{0.916} & \textbf{0.994} & \textbf{1.147} & \textbf{1.837} & \textbf{0.504} & \textbf{0.909} & \textbf{1.074} & \textbf{1.282} & \textbf{1.653} & 2.168\\
& $\pm$ 1e-3 & $\pm$ 1e-3 & $\pm$ 2e-3 & $\pm$ 3e-3 & $\pm$ 3e-3 & $\pm$ 3e-3 & $\pm$ 2e-3 & $\pm$ 2e-3 & $\pm$ 2e-3 & $\pm$ 3e-3 & $\pm$ 3e-3 & $\pm$ 3e-3 & $\pm$ 1e-3 & $\pm$ 1e-3 & $\pm$ 1e-3 & $\pm$ 2e-3 & $\pm$ 2e-3 & $\pm$ 2e-3 & $\pm$ 1e-3 & $\pm$ 2e-3 & $\pm$ 4e-3 & $\pm$ 5e-3 & $\pm$ 6e-3 & $\pm$ 7e-3 \\
\midrule
ERD \cite{frag} & 1.30 & 1.56 & 1.84 & - & 2.00 & 2.38 & 1.66 & 1.93 & 2.28 & - & 2.36 & 2.41 & 2.34 & 2.74 & 3.73 & - & 3.68 & 3.82 & 2.67 & 2.97 & 3.23 & - & 3.47 & 2.92\\
LSTM-3LR \cite{frag} & 1.18 & 1.50 & 1.67 & - & 1.81 & 2.20 & 1.36 & 1.79 & 2.29 & - & 2.49 & 2.82 & 2.05 & 2.34 & 3.10 & - & 3.24 & 3.42 & 2.25 & 2.33 & 2.45 & - & 2.48 & 2.93\\
SRNN \cite{jain} & 1.08 & 1.34 & 1.60 & - & 1.90 & 2.13 & 1.35 & 1.71 & 2.12 & - & 2.28 & 2.58 & 1.90 & 2.30 & 2.90 & - & 3.21 & 3.23 & 1.67 & 2.03 & 2.20 & - & 2.39 & 2.43\\
VGRU-ac & 1.180 & 1.210 & 1.247 & 1.236 & 1.291 & 1.363 & 1.150 & 1.210 & 1.310 & 1.400 & 1.490 & 1.700 & 1.81 & 1.950 & 2.080 & 2.140 & 2.240 & 2.440 & 1.720 & 1.970 & 1.930 & 1.870 & 2.050 & \textbf{2.147}\\
& $\pm$ 3e-4 & $\pm$ 3e-4 & $\pm$ 2e-4 & $\pm$ 3e-4 & $\pm$ 6e-4 & $\pm$ 7e-4 & $\pm$ 2e-4 & $\pm$ 1e-4 & $\pm$ 1e-4 & $\pm$ 1e-4 & $\pm$ 1e-4 & $\pm$ 1e-4 & $\pm$ 1e-4 & $\pm$ 1e-4 & $\pm$ 1e-4 & $\pm$ 1e-4 & $\pm$ 1e-4 & $\pm$ 1e-4 & $\pm$ 1e-4 & $\pm$ 1e-4 & $\pm$ 1e-4 & $\pm$ 1e-4 & $\pm$ 1e-4 & $\pm$ 1e-4 \\
MBR-long \cite{jul} & 0.93 & 1.05 & 1.24 & 1.29 & 1.43 & 1.56 & 1.13 & 1.35 & 1.75 & 1.91 & 2.07 & 2.28 & 1.29 & 2.07 & 2.53 & 2.56 & 2.76 & 3.39 & 1.63 & 2.03 & 2.57 & 2.72 & 2.96 & 2.94\\ 
GRU-d (ours) & 1.311 & 1.333 & 1.369 & 1.364 & 1.350 & 1.370 & 1.275 & 1.305 & 1.386 & 1.466 & 1.530 & 1.702 & 1.943 & 2.062 & 2.201 & 2.255 & 2.342 & 2.486 & 1.744 & 1.980 & 2.026 & 1.994 & 2.214 & 2.172\\
VGRU-d (ours) & 1.108 & 1.146 & 1.211 & 1.200 & 1.220 & 1.280 & 1.090 & 1.160 & 1.240 & 1.330 & 1.370 & 1.560 & 1.670 & 1.800 & 1.940 & 1.980 & 2.060 & 2.320 & 1.749 & 2.037 & 2.011 & 1.868 & 2.088 & 2.318 \\
& $\pm$ 1e-4 & $\pm$ 1e-4 & $\pm$ 2e-4 & $\pm$ 2e-4 & $\pm$ 3e-4 & $\pm$ 2e-4 & $\pm$ 2e-4 & $\pm$ 1e-4 & $\pm$ 1e-4 & $\pm$ 1e-4 & $\pm$ 1e-4 & $\pm$ 1e-4 & $\pm$ 1e-4 & $\pm$ 1e-4 & $\pm$ 1e-4 & $\pm$ 1e-4 & $\pm$ 1e-4 & $\pm$ 1e-4 & $\pm$ 1e-4 & $\pm$ 1e-4 & $\pm$ 1e-4 & $\pm$ 1e-4 & $\pm$ 1e-4 & $\pm$ 1e-4 \\
\bottomrule
\end{tabular}}
\vspace{1mm}
\caption{Long-term motion results: All models are trained on single-action data (SA = single-action). Top set show short-term models including the (MBR-unsup(SA) = Residual unsup. (MA) from \cite{jul} re-trained on SA) and ours sampled for longer duration to match long-term duration. Bottom set shows long-term models by MBR-long = sampling-based loss (SA) from \cite{jul}, ERD and LSTM-3LR from \cite{frag}, SRNN from \cite{jain}), our GRU-d and VGRU-d and VGRU-ac. Since the VTLN-RNN architecture samples from a noise distribution for each forward pass, table shows mean and standard deviation over 30 trials.}
\label{tab:long-term}
\end{table*}
In order to discern the strengths and weaknesses of short-term and long-term models, we computed the NPSS metric on test sequences at 3 different timescales, i.e., 1) short-term: 0-1 s, 2) medium-term: 1-2 s, 3) long-term: 2-4 s, along the prediction timeline for test sequences shown in Table \ref{tab:npss-timescale}. Observe that the short-term models (above the double line) VGRU-r1 (SA) and MBR-unsup (SA) perform competitively with long-term models (below the double line) in the short-term timescale. In the medium-term prediction horizon, the short-term models degrade slightly more than the long-term models, as evidenced by a small gap in the measured NPSS values. However, in the long-term prediction horizon (of 2-4 s), the short-term models degrade significantly relative to the long-term models. This is evidenced by wider gaps in NPSS values. GRU-d and VGRU-d models perform best across all actions and time-horizons, effectively outperforming MBR-long and VGRU-ac. \\
Finally, Table \ref{tab:short-term} shows MSE results for short-term motion prediction experiments on multi-action data on test set sequences. Zero-velocity is a simple, yet  hard-to-beat baseline, introduced in \cite{jul}, which uses the previous frame as the prediction for current one. We can see that VGRU-r1 model is competitive with the state-of-art short-term MBR model as well the quite powerful, zero-velocity baseline. These results show that our proposed VTLN-RNN architecture, augmented with motion-derivative features and our novel multi-objective loss function, can serve as useful predictors of short-term motion prediction as well as powerful long-term motion synthesizers.
Since our models, particularly the long-term ones, contain multiple innovations, we conducted an ablation study to test the utility of each of proposed components: 1) the 2-level processing VTLN-RNN architecture itself (i.e., the backward and forward processing layers), 2) appending a vector of approximate derivatives of joint angles as features, and 3) the multi-objective cost function for parameter optimization. Our ablative study examined multiple variants including: 1) a full VTLN-RNN architecture with derivative features and the multi-objective cost, 2) a full VTLN-RNN architecture with derivative features but without the new cost, 3) a regular RNN model (without the 2-level processing component) with derivative features, and 4) a regular RNN (without the 2-level processing and without appending derivatives as features or use of the proposed cost).  Details of this study are provided in the supplementary. In short, our study revealed that each of these innovations were necessary in improving the performance of the long-term synthesis model, with the cost playing the most important role.

\section{Conclusions and Future Work}
\label{sec:conclusion}
For human motion prediction and synthesis, we introduced the VTLN-RNN architecture, which uses motion derivative features as well as a novel multi-objective loss function to achieve state-of-art performance on long-term motion synthesis. The proposed framework also achieves competitive performance on short-term motion prediction thereby demonstrating general applicability. Furthermore, we proposed a new metric, the Normalized Power Spectrum Similarity (NPSS), and demonstrated that it addresses and alleviates key drawbacks of the commonly-used mean squared error in long-term motion synthesis evaluation. 
Future research directions include incorporating the NPSS metric into the parameter optimization process and developing models that better train on multi-action data across all situations, particularly for long-term motion synthesis. 

\section*{Acknowledgements}
\label{sec:ack}
We gratefully acknowledge partial support from NSF grant CCF$\#$1317560.

\clearpage

{\small
\bibliographystyle{ieee}
\bibliography{egbib}
}

\end{document}